\documentclass[conference]{IEEEtran}
\IEEEoverridecommandlockouts
\usepackage{cite}
\usepackage{amsmath,amssymb,amsfonts}
\usepackage{algorithmic}
\usepackage{graphicx}
\usepackage{textcomp}
\usepackage{xcolor}
\usepackage{amsmath}
\usepackage{hyperref}
\hypersetup{draft}
\def\BibTeX{{\rm B\kern-.05em{\sc i\kern-.025em b}\kern-.08em
    T\kern-.1667em\lower.7ex\hbox{E}\kern-.125emX}}
\begin{document}

\title{Deep Feature Rotation for Multimodal Image Style Transfer}

\author{
\IEEEauthorblockN{Son Truong Nguyen}
\IEEEauthorblockA{
\textit{School of Mechanical Engineering} \\
Hanoi University of Science\\ 
and Technology, Vietnam \\
son.nt185900@sis.hust.edu.vn}
\and 
\IEEEauthorblockN{Nguyen Quang Tuyen}
\IEEEauthorblockA{
\textit{School of Computer Science}\\
\textit{and Engineering} \\
University of Aizu, Japan \\
s1262008@u-aizu.ac.jp}
\and
\IEEEauthorblockN{Nguyen Hong Phuc}
\IEEEauthorblockA{
\textit{Department of Software Engineering} \\
Eastern International University\\ 
Vietnam (Corresponding author)\\
phuc.nguyenhong@eiu.edu.vn}
}

\maketitle

\begin{abstract}
Recently, style transfer is a research area that attracts a lot of attention, which transfers the style of an image onto a content target. Extensive research on style transfer has aimed at speeding up processing or generating high-quality stylized images. Most approaches only produce an output from a content and style image pair, while a few others use complex architectures and can only produce a certain number of outputs. In this paper, we propose a simple method for representing style features in many ways called Deep Feature Rotation (\textit{DFR}), while not only producing diverse outputs but also still achieving effective stylization compared to more complex methods. Our approach is representative of the many ways of augmentation for intermediate feature embedding without consuming too much computational expense. We also analyze our method by visualizing output in different rotation weights. Our code is available at \url{https://github.com/sonnguyen129/deep-feature-rotation}.
\end{abstract}

\begin{IEEEkeywords}
Neural style transfer, transfer learning, deep feature rotation
\end{IEEEkeywords}

\section{Introduction}
Style transfer aims to re-render a content image $I_{c}$ by using the style of a different reference image $I_{s}$, which is widely used in computer-aid art generation. The seminal work of Gatys \textit{et al.} \cite{b1} showed that the correlation between features encoded by a pretrained deep convolutional neural network \cite{b9} can capture the style patterns well. However, stylizations are generated on an optimization scheme that is prohibitively slow, which limits its practical application. This time-consuming optimization technique has prompted scientists to look into more efficient methods. Many neural style transfer methods use feed-foward networks \cite{b2,b3,b4,b5,b6,b7,b8,b13} to synthesize the stylization. Besides, the universal style transfer methods \cite{b5,b6,b7,b8} inherently assume that the style can be represented by the global statistics of deep features such as Gram matrix \cite{b1}. Since the learned model can only synthesize for one specific style, this method and the following works \cite{b11,b12,b4,b17,b13,b14,b15,b16} are known as Per-Style-Per-Model method. Furthermore, these works \cite{b18,b19,b20,b21} are categorized to Multiple-Style-Per-Model method and Arbitrary-Style-Per-Model \cite{b5,b3,b8,b22,b23,b6,b24,b25,b10,b26,b28,b30}. These findings address some of the style transfer issues, such as balancing content structure and style patterns while maintaining global and local style patterns. Unfortunately, while the preceding solutions improved efficiency, they failed to transfer style in various ways since they could only produce a single output from a pair of content and style images.

Recently, \cite{b11} introduced a multimodal style transfer method called DeCorMST that can generate a set of output images from the pair of content and style images. Looking deeper into that method, it can be easily noticed that it tries to generate multiple outputs by optimizing the output images using multiple loss functions based on correlation functions between feature maps, such as gram matrix, Pearson correlation, covariance matrix, euclidean distance, and cosine similarity. This aided the procedure in producing five outputs that were equivalent to the five loss functions utilized, but the five outputs did not represent the diversity of style transfer. Simultaneously, this method slows down the back-propagation process, making it difficult to use in reality.

Facing the aforementioned challenges, we propose a novel deep feature augmentation method that aims to rotate features at different angles named \textit{\textbf{D}eep \textbf{F}eature \textbf{R}otation} (\textit{\textbf{DFR}}). Our underlying motivation is to create new features from the features extracted from well-trained VGG19 \cite{b9}, thereby observing the change in stylizations. In the meanwhile, there are many ways to transform intermediate features in style transfer. However, their works produce only one or few outputs from a content and style image pair. Our method is possible to generate infinite outputs with a simple approach.

Our main contributions are summarized as follows:
\begin{itemize}
    \item We qualitatively analyze the synthesized outputs of different feature rotating.
    \item We propose a novel end-to-end network that produces varieties of style representations, each representing a particular style pattern based on different angles.
    \item The introduced rotation weight indicator clearly shows the trade-off between the original feature and the feature after rotation.
\end{itemize}

\section{Related Work}

\begin{figure*}[t]
    \centering
    \includegraphics[width=\textwidth]{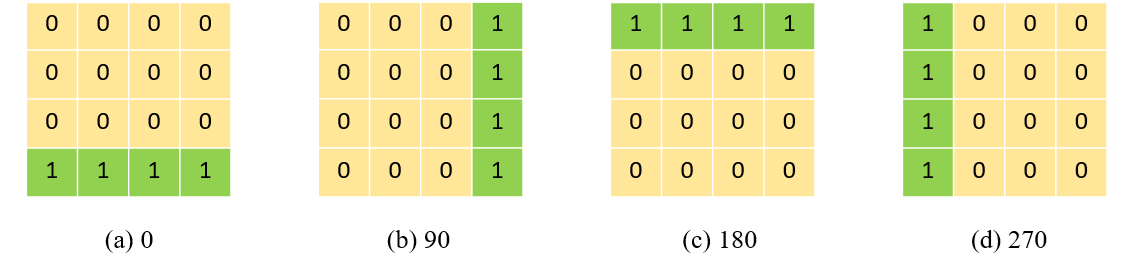}
    \caption{Illustration of rotating mechanism at different degrees. From (a) to (d) illustrate the rotation mechanism equivalent to the angles 0, 90, 180, 270, respectively. We feed both feature maps to the Rotation module, which rotates all feature maps in all two dimensions after encoding the content and style images. The illustration shows that our method works on feature maps, so this change will produce differences in the output image.}
    \label{fig}
\end{figure*}

\begin{figure*}[t]
    \centering
    \includegraphics[width=\textwidth]{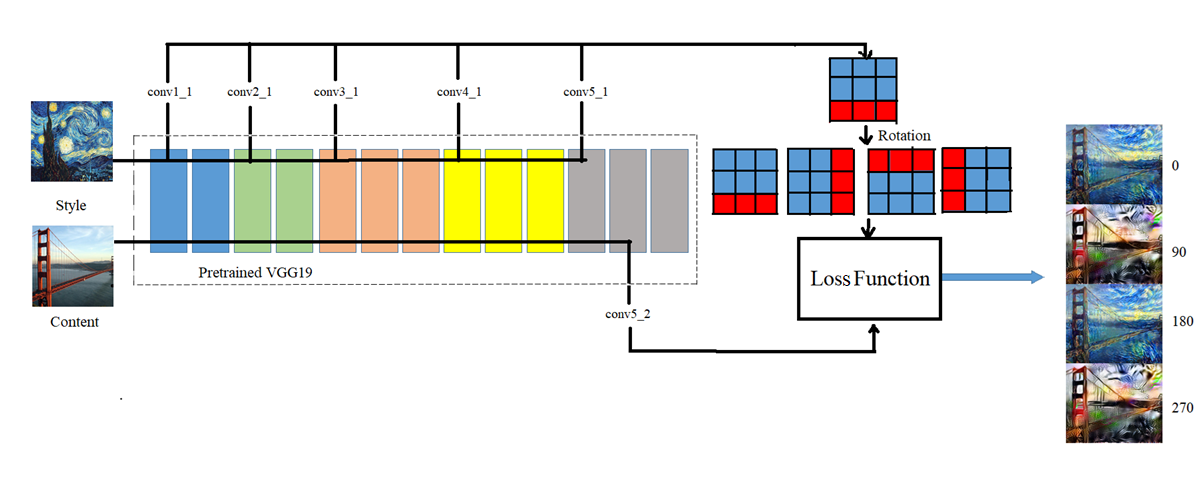}
    \caption{Style transfer with deep feature rotation. The architecture consists of a backbone of VGG19 extracting features from style and content images. These feature maps computed by the loss function, will produce the output depending on the rotation. With each rotation, the model will get a different result. The model optimizes the loss function by standard error back-propagation. The generating efficiency is an outstanding performance, entirely possible to balance the content structure and the style patterns.}
    \label{fig}
\end{figure*}

Recent neural style transfer methods can be divided into two categories: Gram-based methods and patch-based methods. The common idea of the former is to apply feature modification globally. Huang \textit{et al.} \cite{b5} introduced AdaIN that aligns channel-wise mean and variance of a style image to a content image by transferring global feature statistics. Jing \textit{et al.} \cite{b23} improved this method by dynamic instance normalization,  in which weights for intermediate convolution blocks are created by another network with the style picture as input. Li \textit{et al.} \cite{b8} proposed to learning a linear transformation according to content and style features by using a feed-forward network. Singh \textit{et al.} \cite{b10} combined self-attention and normalization called SAFIN in a unique way to mitigate the issues with previous techniques. WCT \cite{b6} performs a pair of transform process with the co-variance instead of variance, whitening and coloring, for feature embedding within a pretrained encoder-decoder module. Despite the fact that these methods successfully complete the overall neural style transfer task and make significant progress in this field, local style transfer performance is generally unsatisfactory due to the global transformations they employ making it difficult to account for detailed local information.

Chen and Schmidt \cite{b3} introduced the first patch-based approach called Style Swap, which matched each content patch to the most similar style patch and swapped them on normalized cross-correlation measure. CNNMRF \cite{b29} enforced local patterns in deep feature space based on Markov random fields. Another patch-based feature decoration method presented by Avatar-Net \cite{b7} converts content features to semantically nearby style features while minimizing the gap between their holistic feature distributions. That method combines the idea of style swap and AdaIN module. However, these methods could not synthesize high-quality outputs when content and style targets have similar structures. 

In recent years, Zhang \textit{et al.} \cite{b30} developed the multimodal style transfer, that method that seeks for a sweet spot between Gram-based and patch-based methods. For the latter, self-attention mechanism in style transfer gives outstanding performance in assigning different style patterns to different regions in an image. SANet \cite{b28}, SAFIN \cite{b10}, AdaAttN \cite{b26} are examples. SANet \cite{b28} firstly introduced the first attention method for feature embedding in style transfer. SAFIN \cite{b10} extended SANet \cite{b28} by using a self-attention module to learn how to create the parameters for our spatially adaptive normalization module, which makes normalization more spatially flexible and semantically aware. AdaAttN \cite{b26} calculated attention scores using both shallow and deep features, as well as properly normalizing content features so that feature statistics are well aligned with attention-weighted mean and variance maps of style features on a per-point basis. As a result, style transfer is more adaptable and has received more attention from academic, industrial, and art communities in several years.  

Most of the above approaches have different feature transformations such as normalization \cite{b5,b23,b8,b10}, attention \cite{b28,b10,b26} or WCT \cite{b6}, etc.. However, our method will focus on exploiting feature transformations by rotating by $90^{\circ}, 180^{\circ}, 270^{\circ}$. Many other simple transformations can be exploited in the future.

\begin{figure*}[t]
    \centering
    \includegraphics[width=\textwidth]{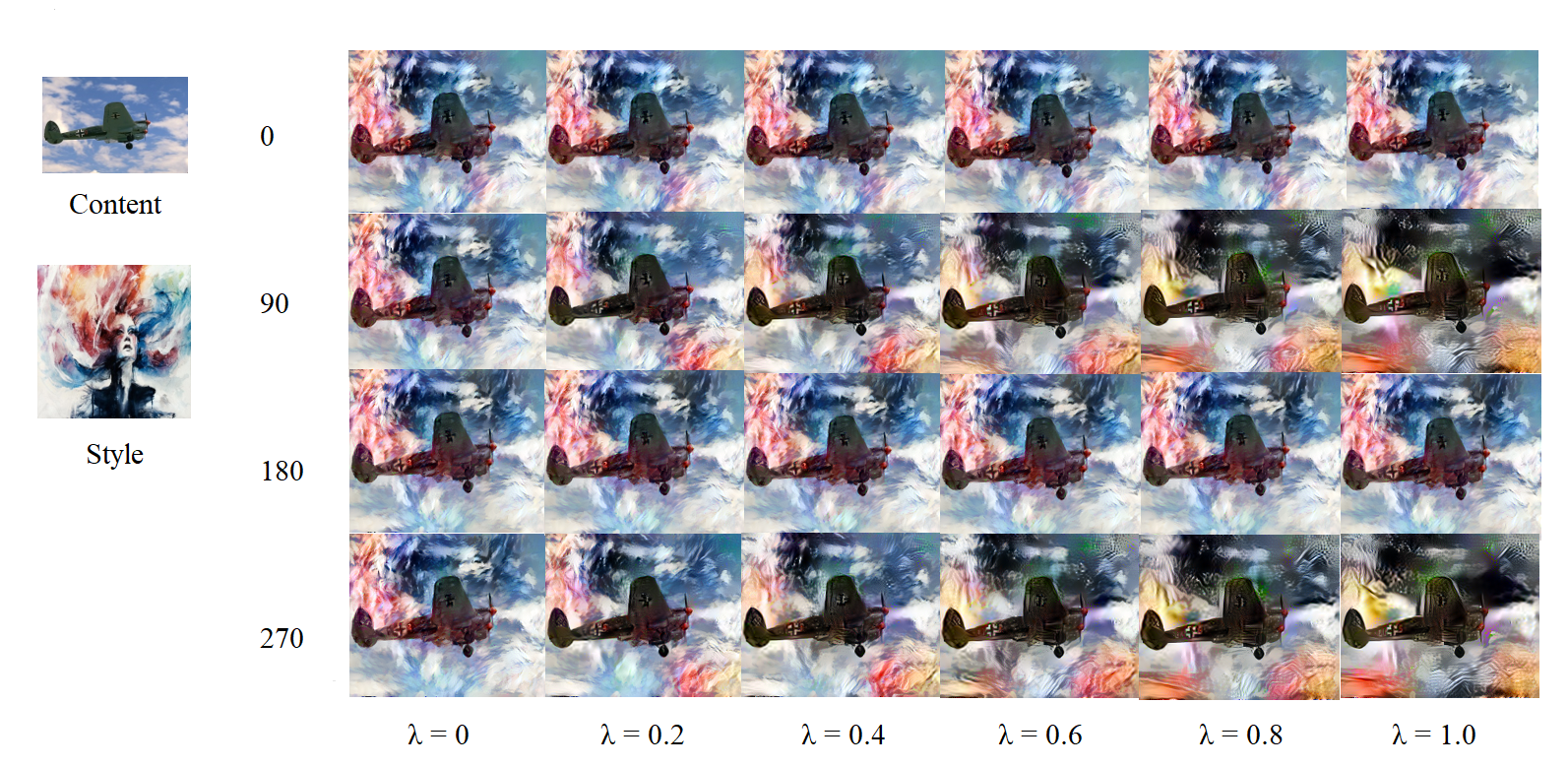}
    \caption{Stylization in different rotation weights. While rows indicate the results corresponding to each rotated angles $0^{\circ}, 90^{\circ}, 180^{\circ}, 270^{\circ}$, respectively, the columns show rotation weight $\lambda$ 0, 0.2, 0.4, 0.6, 0.8, 1.0 respectively. The near rotation weight values also don't result in a lot of variation in the outputs; the larger rotation weight, the more visible the new textures will be.}
    \label{fig}
\end{figure*}

\section{Proposed Method}

\subsection{Background}

Gatys \textit{et al.} introduced the first algorithm \cite{b1} that worked well for the task of style transfer. In this algorithm, a VGG19 architecture \cite{b9} pretrained on ImageNet \cite{b31} is used to extract image content from an arbitrary photograph and some appearance information from the given well-known artwork.

Given a color image $x_{0} \in \mathcal{R}^{W_{0} \times H_{0} \times 3}$, where $W_{0}$ and $H_{0}$ are the image width and height. The VGG19 is capable of reconstructing representations from intermediate layers, which encode $x_{0}$ into a set of feature maps $\{F^l(x_{0}\}^L_{l=1}$, where $F^l : \mathcal{R}^{W_{0} \times H_{0} \times 3} \rightarrow \mathcal{R}^{W_{l} \times H_{l} \times D_{l}}$ is the
mapping from the image to the tensor of activations of the $l^{th}$ layer, which has $D_{l}$ channels of spatial dimensions $W_{l} \times H_{l}$. The activation tensor $F^l(x_{0})$ can be stored into a matrix $F^l(x_{0}) \in \mathcal{R}^{D_{l} \times M_{l}}$, where $M_{l} = W_{l} \times H_{l}$. Style information captured by computing the correlation between activation channels  $F_{l}$ of layer $l$. . These feature correlations are given by the Gram matrix $\{G^l\}^L_{l = 1}$ where $G^l \in \mathcal{R}^{D_{l} \times D_{l}}$:
$$[G^l(F^l)]_{ij} = \sum\limits_{k}^{} F^l_{ik}F^l_{jk}$$.

Given a content image $x^c_{0}$ and a style image $x^s_{0}$, Gatys built the content components of the newly stylised image by penalising the difference of high-level representations derived from content and stylised images, and further build the style component by matching Gram-based summary statistics of style and stylised images. An image $x^*$ that presents the content of stylization is synthesized by solving 
$$ x^* = \operatorname*{argmin}_{x_{0} \in \mathcal{R}^{W_{0} \times H_{0} \times 3}} \alpha L_{content}(x^c_{0},x) + \beta L_{style}(x^s_{0},x)$$
with
$$L_{content} (x^c_{0},x) = \frac{1}{2}  \|F^l(x) - F^l(x_{0})\|^2_{2}.$$
$$L_{style} (x^s_{0},x) = \sum\limits_{l=1}^{L} \frac{w_{l}}{4D^2_{l}M^2_{l}} \|G^l(F^l(x)) - G^l(F^l(x_{0}))\|^2_{2}.$$
where $w_{l} \in \{0,1\}$ are the weighting factors of the contribution of each layer to the total loss,  $\alpha$ and $\beta$ are the weighting factors for the content and style reconstruction, respectively.

\subsection{Deep Feature Rotation}

Our method differs from Gatys' method in that we can generate multiple outputs and it is a simple approach compared to many other methods. After encoding the content and style image, we will get a set of feature maps. We can rotate to many different angles, but in the framework of paper, I only rotate to 90, 180, 270 degrees (See illustration in Fig. 1). Noted that rotation is only performed on feature maps, other dimensions will remain unchanged. We denote it as $W_{(n,c,h,w)} \in  \mathcal{R}^{n \times c \times h \times w}$ and feature maps after rotating as $W_{r}$. The above rotating process can be formulated as follows.

\begin{itemize}
\item With 90 degrees:
\begin{align*}
    W_{r} &= W_{i,j,q,k}, \forall \ 0 \leq i \leq w, \\
    &0 \leq j \leq h, 0 \leq q \leq c, 0 \leq k \leq n 
\end{align*}

\item With 180 degrees: 
\begin{align*}
    W_{r} = W_{i,j,q::-1,k}, \forall \ 0 \leq i \leq n, \\
    0 \leq j \leq c, 0 \leq q \leq h, 0 \leq k \leq w
\end{align*}

\item With 270 degrees:
\begin{align*}
    W_{r}= W_{i,j,q::-1,k}, \forall \ 0 \leq i \leq w, \\
    0 \leq j \leq h, 0 \leq q \leq c, 0 \leq k \leq n
\end{align*}
\end{itemize} 

\begin{figure*}[t]
    \centering
    \includegraphics[width=\textwidth]{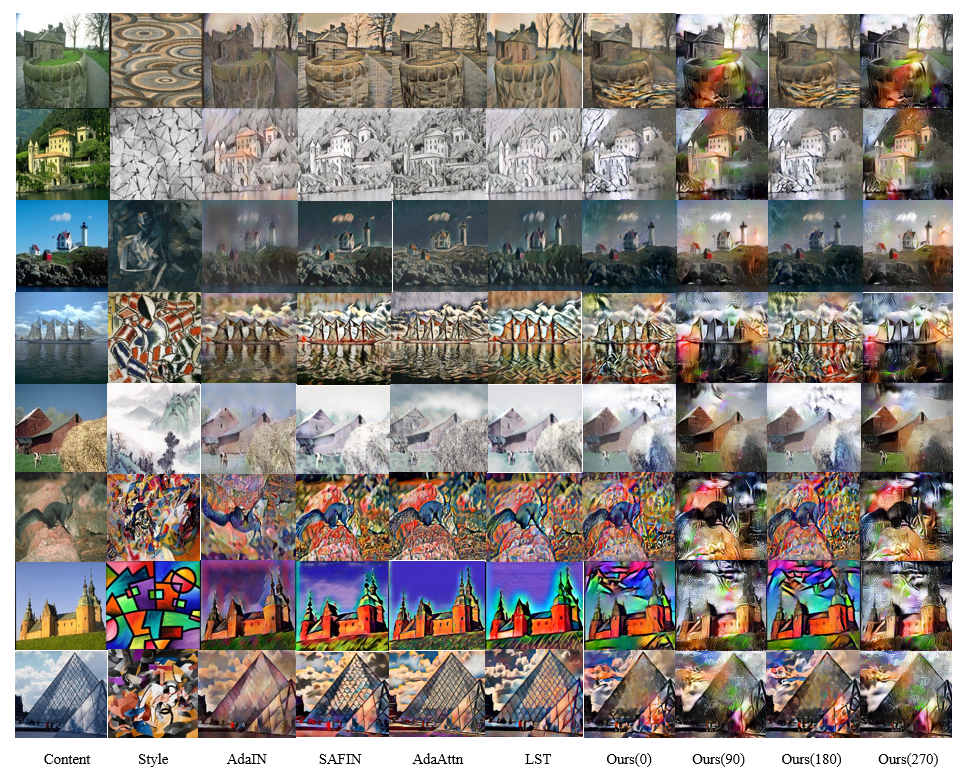}
    \caption{Comparison with other state-of-the-art methods in style transfer. From left to right, the content image, the style image, the AdaIN \cite{b5}, SAFIN \cite{b10}, AdaAttN \cite{b26}, LST\cite{b8} methods, and our results are based on the rotation $0^{\circ}, 90^{\circ}, 180^{\circ}, 270^{\circ}$. Our results are highly effective in balancing global and local patterns, better than AdaIN and LST, but slightly worse than SAFIN and AdaAttn. The obvious difference between our method and SAFIN, AdaAttn is that the sky patterns in the image, our results tend to generate darker textures.}
    \label{fig}
\end{figure*}

Rotating the feature to different angles plays a role in helping the model not only learn a variety of features, but also cost less computationally expensive compared to other methods. Similar to the above three rotation angles, creating features in other rotation angles is completely simple because there is no need to care about statistical values or complicated calculation steps. 

To see the relationship between the original feature maps and the rotated feature maps, we introduce rotation weight $\lambda$ so the final feature map is defined as follows:
$$\hat{W_{r}} = (1 - \lambda)W + \lambda W_{r}$$

The trade-off between the original feature maps and the rotated feature maps is represented by rotation weight. When $\lambda = 0$, the network tries to recreate the baseline output as closely as possible, and when $\lambda = 1$, it seeks to synthesize the most stylized image. By moving from 0 to 1, as illustrated in Fig. 3, a seamless transition between baseline-similarity and rotate-similarity may be detected.

After applying rotation weight, we received four different sets of output feature maps corresponding to rotated angles $0^{\circ}, 90^{\circ}, 180^{\circ}, 270^{\circ},$ respectively(Fig. 2). Then, we will calculate four different loss functions $L$ with four sets of feature map $\hat{W_{r}}$ as follows:
$$ L(x) = \alpha L_{content}(x) + \beta L_{style}(x)$$
with
$$L_{content} (x) = \frac{1}{2}  \|F^l(x) - \hat{W_{r}}\|^2_{2}.$$
$$L_{style} (x) = \sum\limits_{l=1}^{L} \frac{w_{l}}{4D^2_{l}M^2_{l}} \|G^l(F^l(x)) - G^l(\hat{W_{r}})\|^2_{2}.$$

We choose content weight $\alpha$, style weight $\beta$ of $10^4$, 0.01, respectively. Using standard error back-propagation, the optimization method will minimize the loss balancing the output image's fidelity to the input content and style image. 
      
Finally, we received four output images. This is also the special feature of the model compared to previous approaches. From a pair of content and style images, it is possible to generate an infinite number of output images. The generated images retain the global and local information effectively. In Fig.3, the image pairs 0 - 180, 90 - 270 degrees have many similar patterns, the reason is that the feature map is symmetrical. The close rotation weight values also don't cause any significant difference between the outputs, the larger the rotation weight will produce more visible the new textures. The resulting new method can be applied for other complex methods in style transfer.

\section{Experiments}

\subsection{Implementing Details}

Similar to Gatys' method, we choose a content image and a style image to train our method. $\lambda$ is selected with the values: 0, 0.2, 0.4, 0.6, 0.8, 1.0, respectively. \textit{Adam} \cite{b32} with $\alpha$, $\beta_{1}$, and $\beta_{2}$ of 0.002, 0.9, and 0.999, is used as solver. We resize all image to $412 \times 522$. The training phase lasts for 3000 iterations on approximate 400 seconds for each pair of content and style image. 

\subsection{Comparison with State-of-the-Art Methods}

\textbf{Qualitative Comparisons.} As shown in Fig. 4, we compare our method with state-of-the-art style transfer methods, including AdaIN \cite{b5}, SAFIN \cite{b10}, AdaAttN \cite{b28}, LST \cite{b8}. AdaIN \cite{b5} directly transfers second-order statistics of content features globally so that style patterns are transferred with severe content leak problems ($2^{nd}, 3^{rd}, 4^{th}$, and $5^{th}$ rows). Besides, LST \cite{b8} changed features using linear projection, resulting in relatively clean stylization outputs. SAFIN \cite{b10}, AdaAttN \cite{b28} show high efficiency of self-attention mechanism in style transfer. All of them achieve a better balance between style transferring and content structure-preserving. Although the stylization performance has not been as effective as attention-based methods, our method shows a variety in style transfer. In the case of $90^{\circ}$ and $270^{\circ}$ degrees ($8^{th}$ and $10^{th}$ columns), new textures appeared in the corners of the output image. This will help to avoid being constrained to a single output.

\begin{figure}[ht]
    \centering
    \includegraphics[width=0.5\textwidth]{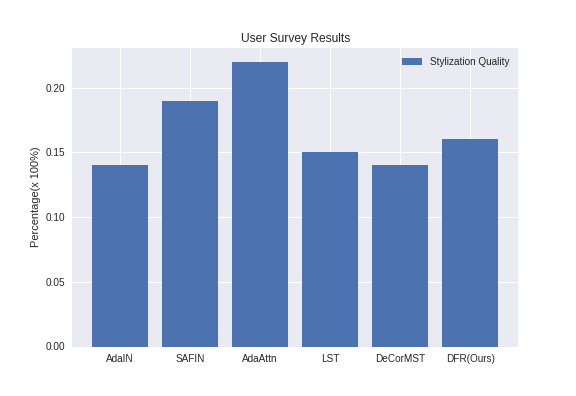}
    \caption{\textbf{User Study Results}. Percentage of the votes that each method received.}
    \label{fig:user}
\end{figure}

\textbf{User Study.} We undertake a user study comparable to \cite{b10} to dig deeper into the six approaches: AdaIN \cite{b5}, SAFIN \cite{b10}, AdaAttN \cite{b26}, LST \cite{b8}, DeCorMST \cite{b11}, and DFR. We use 15 content images from MSCOCO dataset \cite{b35}, and 20 style images from WikiArt \cite{b36}. Using the disclosed codes and default settings, we generate 300 results for each technique. 20 content-style pairs are picked at random for each user. For each style-content pair, we present the styled outcomes of 6 ways on a web page in random order. Each user is given the opportunity to vote for the option that he or she likes the most. Finally, we collect 600 votes from 30 people to determine which strategy obtained the most percentage of votes. As illustrated in Fig. 5, the stylization quality of DFR is slightly better than AdaIN, LST and worse than the attention-based method. This user study result is consistent with the visual comparisons (in Fig. 4).

\begin{table}
    \centering
    \caption{ Running time comparison between multimodal methods}
    \begin{tabular}{|c|c|c|c|c|c|c|}
        \hline
        Method & DeCorMST & DFR-1 & DFR-2 & DFR-3 & DFR-4 \\
        \hline
        Time & 3 days & 361s & 377s & 396s & 417s \\
        \hline
    \end{tabular}
\end{table}

\textbf{Efficiency.}  We further compare the running time of our
methods with multimodal ones \cite{b11}. Tab. 1 gives
the average time of each method on 30 image pairs with size of 412 × 522. Our DFR with different X (DFR-X) equivalent to the number of outputs generated. It can be seen that the speed of DFR is much faster than DeCorMST. As the number of outputs increases, the model inference time also increases, but not too much. We will look to reduce the inference time in the future.

\section{Conclusion}

In this work, we proposed a new style transfer algorithm that transforms intermediate features by rotating them at different angles. Unlike the previous methods that only produce one or little output from a content and style image pair, our proposed method can produce a variety of outputs effectively and efficiently. Furthermore, the method only rotates the representative by four angles, with different angles will give more special results. Experimental results demonstrate that we can improve our method in many ways in the future. In addition, applying style transfer to images that are captured from different cameras \cite{b21} is challenging and promising, since constraints from these images need to be preserved.

\end{document}